# Enhancing Convergence Speed with Feature-Enforcing Physics-Informed Neural Networks: Utilizing Boundary Conditions as Prior Knowledge for Faster Convergence

Mahyar Jahaninasab[a], Mohamad Ali Bijarchi [a,*]

[a] Department of Mechanical Engineering, Sharif University of Technology, Tehran, Iran[1].

*Corresponding author: bijarchi@sharif.edu

**Abstract:**

This study introduces an accelerated training method for Vanilla Physics-Informed-Neural-Networks (PINN) addressing three factors that imbalance the loss function: initial weight state of a neural network, domain to boundary points ratio, and loss weighting factor. We propose a novel two-stage training method. During the initial stage, we create a unique loss function using a subset of boundary conditions and partial differential equation terms. Furthermore, we introduce preprocessing procedures that aim to decrease the variance during initialization and choose domain points according to the initial weight state of various neural networks. The second phase resembles Vanilla-PINN training, but a portion of the random weights are substituted with weights from the first phase. This implies that the neural network's structure is designed to prioritize the boundary conditions, subsequently affecting the overall convergence. Three benchmarks are utilized: two-dimensional flow over a cylinder, an inverse problem of inlet velocity determination, and the Burger equation. It is found that incorporating weights generated in the first training phase into the structure of a neural network neutralizes the effects of imbalance factors. For instance, in the first benchmark, as a result of our process, the second phase of training is balanced across a wide range of ratios and is not affected by the initial state of weights, while the Vanilla-PINN failed to converge in most cases. Lastly, the initial training process not only eliminates the need for hyperparameter tuning to balance the loss function, but it also outperforms the Vanilla-PINN in terms of speed.

## 1. Introduction

The rapid advancements in the field of artificial intelligence (AI) have inspired researchers to explore new ways to incorporate AI techniques into their respective fields. Raissi et al.'s pioneering work demonstrated the potential of neural networks as a powerful tool for solving partial differential equations (PDEs) [1]. The incorporation of a PDE loss term into the loss function, along with the mean squared error of predicting boundary conditions, led to the achievement of this solution. This breakthrough has since motivated many researchers to further investigate the use of deep learning techniques in various fields, including physics, finance, and engineering [2-4]. The vanilla form of PDEs solver, known as Physics-Informed Neural Networks (PINNs), represents a novel technique that has shifted attention from data-driven models to this emerging field [5-8]. The Vanilla PINN possesses remarkable versatility due to its mesh-free characteristics, enabling it to tackle a diverse range of challenges. It has proven its efficacy in different scenarios, ranging from forward to inverse problems [9]. For instance, it has successfully addressed the lid-driven cavity test case, which is governed by the incompressible Navier-Stokes equation [10]. Additionally, it has demonstrated its ability in handling multiphase problems [11], as well as scenarios involving flow past a cylinder and conjugate heat transfer [12]. These accomplishments solidify its status as a dependable approach for addressing various fluid mechanics dilemmas. However, the convergence issues and slow training time remain challenges. As such, the quest to enhance the convergence speed of PINN has become a prominent research topic, marking the beginning of a new era in the field. An enhanced, yet more expensive approach to speeding up the convergence speed of PINNs is through the use of decomposition methods. One example of using this technique to speed up the convergence of PINNs is the work of Alena Kopaničáková et al. They employed a decomposed neural network strategy by dividing it into sub-networks, with each sub-network being trained separately on a dedicated GPU [13]. While techniques such as decomposition methods can lead to a faster convergence rate, they require expensive resources for solving PDEs. This represents a potential drawback, even for simple problems. However, there are other approaches that can accelerate convergence speed without the need for such resources. One such approach is transfer learning [14]. Transfer learning is a powerful approach that entails leveraging knowledge from a pre-trained model to enhance the learning of another model [15]. Although recent advancements in transfer learning techniques have demonstrated their potential in training deep learning models [16-18], they are not widely applied to solve PDEs using neural networks. The fact that solutions to PDEs are specific to the problem being solved can make it difficult to effectively use transfer learning. The transferability of such tasks is considerably more challenging than that of classic tasks like image classification. Despite this difficulty, researchers have discovered novel applications of transfer learning for PINNs [19-21]. In addition to transfer learning, another technique that can help accelerate the convergence of neural networks is warm-up training. Warm-up training can help the model slowly adapt to the data and allows adaptive optimizers to compute correct statistics of the gradients [22]. Few studies have investigated the potential of warm-up training in this field [23]. Junjun Yan et al. utilized warm-up training to generate pseudo labels, which were subsequently employed in the main training loop [24]. Another popular approach to warm-up training involves initial training for a defined number of iterations with the ADAM optimizer, followed by another training loop with LBFGS, which leads to faster convergence in comparison to vanilla PINN [25].

Another field of study aimed at improving the convergence speed of vanilla PINN involves addressing the inherent problems with this method. As research in this area has progressed, new and improved methods have been developed that build upon the original vanilla form. These methods involve modifications to various components of the vanilla form, such as the architecture [26], activation function [27], training method [28], sampling [29], and loss function [30]. One of the challenges in PINN is dealing with an imbalanced loss function, where the PDE loss term is more frequent than the boundary loss term. This issue can lead to a biased model that converges poorly. Lu et al. presented DeepXDE as a means to optimize the loss function of PINNs for PDEs. They proposed an approach called residual-based adaptive refinement (RAR), which incorporates supplementary collection points into areas characterized by high PDE residuals [31]. Nabian et al. introduced a collection point resampling strategy that utilizes importance sampling, relying on the loss function distribution to improve convergence [32]. Several studies have concentrated on addressing training issues through the use of adaptive sampling strategies [33-36]. These studies aim to optimize the training process by strategically selecting the most effective domain points at each stage of training. The primary objective of this approach is to maintain a balanced training process, which in turn, facilitates a more efficient convergence process. By ensuring a balanced distribution of domain points, these methods aim to optimize the learning process and improve the overall performance of the model. An alternative approach to selecting the optimal domain points at each training step involves the application of loss weighting strategies, which allow for direct control over the value of each term in the loss function. Wang et.al introduced a method known as 'learning rate annealing for PINNS' that dynamically adjusts the weights during the training process [40]. The fundamental principle of this method is the automatic adjustment of weights based on the statistics of the back-propagated gradient during model training, ensuring a balanced distribution across all elements of the loss function. This approach proved to be a more effective alternative to the previously employed method of manually adjusting the weights to balance the loss function [41]. In a different research study, Maddu et. al introduced a method that utilizes gradient variance to achieve a balanced training process for PINNs. This method, known as Inverse-Dirichlet Weighting, also incorporates a momentum update with a specific parameter. Experimental results indicate that this approach effectively mitigates the issue of gradient vanishing [42]. However, these methods, which use adaptive weighting schemes and dynamic updates for loss function balancing, may not be universally effective. The process of determining optimal weight values can be computationally intensive and potentially unsuccessful. The existing studies [33-42] primarily concentrate on one factor to balance the loss function, neglecting other potential factors. These studies do not take into account all the elements that contribute to balancing the loss function. Furthermore, their proposed methods require calculation at every step of training, which makes them computationally expensive. This focus on a single aspect and the computational cost associated with their methods highlight the need for more comprehensive and efficient approaches.

In this study, we introduce additional factors that could cause an imbalance in the loss function. We propose a novel two-phase training methodology that balances the loss function by taking all these factors into account. In the first phase of our methodology, we introduce a new loss function that is more efficient than other balancing strategies, as it comprehensively addresses all factors contributing to the balance of the loss function. We further enhance this phase by hypothesizing that an initial weight state with lower variance is beneficial, and by selecting the input space based on the Xavier initialization scheme. The second phase of our methodology is similar to the training of vanilla Physics-Informed Neural Networks (PINN). However, a key distinction lies in the utilization of the weights produced in the first phase. These weights are used to replace a proportion of the random weights in the neural network structure. In our methodology, the strategic replacement of weights not only balances the training process of the vanilla PINN for all factors but also offers a computational advantage as it is faster than traditional loss weighting strategies and adaptive sampling methods. In this study, we employed three benchmarks to evaluate our methodology against the vanilla PINNs. Included benchmarks are as follows: a two-dimensional (2D) flow over a cylinder, an inverse problem to determining the inlet velocity in a 2D flow over a cylinder, and the Burgers' equation. Notably, in Section 2.1, we present the formulation of the vanilla PINN loss function as applied to the benchmarks under consideration, and we discuss the training challenges raised with this formulation in Section 2.2. In the subsequent sections, we introduce our alternative approach and conduct a comparative analysis with vanilla PINN. The code for this study is publicly accessible and can be found at the following GitHub repository: https://github.com/mahyar-jahaninasab/Feature-Enforcing-PINN

**2.1 Vanilla PINN Loss Function**

In vanilla PINN, the loss function is defined by equation (1). The typical process for forming this loss function involves calculating derivatives based on the input for the first term while incorporating all boundary conditions into the second term. Please note that in the studies mentioned earlier [40-42], $\lambda$ is used as the weighting factor in conjunction with adaptive strategies.

$$Loss\ Function\ =\ L_{pde}\ +\ \lambda \times L_{boundary} \tag{1}$$

Navier-Stokes equations have been the primary choice for addressing laminar flow in the majority of research studies in this field [37-39]. Navier-Stokes equations are used to address the first two benchmarks. In these two benchmarks $L_{pde}$ is defined as follows:

$$L_{pde} = f_0^2 + f_1^2 + f_2^2 \tag{2}$$

$$f_0 = \frac{\partial u}{\partial x} + \frac{\partial v}{\partial y} \tag{3}$$

$$f_1 = \rho \left( u \times \frac{\partial u}{\partial x} + v \times \frac{\partial u}{\partial x} \right) - \frac{\partial \sigma_{xx}}{\partial x} - \frac{\partial \sigma_{xy}}{\partial y} \tag{4}$$

$$f_2 = \rho \left( u \times \frac{\partial v}{\partial x} + v \times \frac{\partial v}{\partial y} \right) - \frac{\partial \sigma_{xy}}{\partial x} - \frac{\partial \sigma_{yy}}{\partial y} \tag{5}$$

$$\sigma_{xx} = -p + 2 \times \mu \times \frac{\partial u}{\partial x} \tag{6}$$

$$\sigma_{yy} = -p + 2 \times \mu \times \frac{\partial v}{\partial y} \tag{7}$$

$$\sigma_{xy} = \mu \left( \frac{\partial u}{\partial y} + \frac{\partial v}{\partial x} \right) \tag{8}$$

Where $\rho$ represents the fluid density, $u$ and $v$ denote the $x$ and $y$ components of the velocity vector, $\mu$ represents the dynamic viscosity, $p$ corresponds to the fluid pressure, and $\sigma$ denotes the stress. The loss function quantifies the difference between the predicted output of the neural network and the boundary value and uses PDE loss to understand the governing equation. In the first benchmark, the density of the fluid is $1 \frac{kg}{m^3}$ and the viscosity of the fluid is $0.02 \frac{kg}{m.sec}$ with an inlet Reynolds number of 26.6. The channel in this benchmark is characterized by a length of 1m and a height of 0.4m. Within this channel, the 2D cylinder has a diameter of 0.1m. The positioning of the cylinder is such that it is located 0.15m from the inlet and an equivalent distance from the bottom of the channel. The boundary condition applied to the wall of the channel is the no-slip condition, implying that the fluid velocity at the wall is zero. Furthermore, the pressure at the outlet of the channel is defined as zero. The velocity at the inlet is defined as follows:

$$u = \frac{4 \times y \times (0.4 - y)}{0.4^2} \tag{9}$$

The first benchmark is a forward problem where known boundary conditions are used to calculate velocity and pressure in the domain. On the other hand, the second benchmark presents an inverse problem. In this case, the inlet boundary condition is not included. Instead, the loss function utilizes 60 domain points with known velocity values. The final goal is to determine the inlet velocity.

$$L_{boundary} = \|Predictions(x, y \,|w) - Labels\|^2 \tag{10}$$

Equation (11) quantifies the difference between the predicted output of the neural network and the boundary condition or the data available for the inverse problem. The formulation of this new loss term, which is central to the solution of the inverse problem, is defined as follows:

$$L_{Inverse\ problem} = \frac{1}{n}\sum_{i=0}^{n} \|Predictions(x, y\,|w) - u(x,y)\|^2 + \|Predictions(x, y\,|w) - v(x,y)\|^2 \tag{11}$$

The final benchmark, known as Burger's equation, is an unsteady problem with $\mu = 0.01 \frac{kg}{m.sec}$ (Equation (12)). In this benchmark, Burger's equation is solved within a specific domain and time frame. The spatial variable, denoted as x, spans from 0 to 4 meters. Concurrently, the time interval for the solution is defined from 0 to 5 seconds. The initial condition for this problem is derived by substituting $u(x, t = 0)$ into Equation (13). The boundary conditions are defined such that the value of the output is zero when x equals either 0 or 4. With these settings in place, Burger's equation can be solved to yield an exact solution which is Equation (13).

$$f_0 = +\frac{\partial u}{\partial t} + u\frac{\partial u}{\partial x} - \mu\frac{\partial^2 u}{\partial x^2} \tag{12}$$

$$u(x,t) = \frac{2 \times 0.01 \times \pi \times \sin(\pi x) \times e^{-0.01\pi^2 \times (t-5)}}{2 + \cos(\pi x) \times e^{-0.01\pi^2 \times (t-5)}} \tag{13}$$

$L_{pde}$ for the last benchmark is formed by calculating the square of $f_0$ and taking the mean across the entire domain point. In conclusion, based on the PDE and the boundary conditions, the loss function for each benchmark is formed as follows:

$$\text{Loss Function} = L_{pde} + \lambda \times (no-slip \text{ boundary condition on side walls} + \text{ inlet velocity boundary condition} + \text{ no} - slip \text{ boundary condition on the cylinder} + \text{outlet pressure}) \tag{14}$$

$$\text{Loss Function} = L_{pde} + \lambda \times (\text{ no} - slip \text{ boundary condition on side walls} + L_{Inverse\ problem} + \text{no} - slip \text{ boundary condition on the cylinder} + \text{outlet pressure}) \tag{15}$$

$$\text{Loss Function} = L_{pde} + \lambda \times (u(x,t=0) + u(x=0,t) + u(x=4,t)) \tag{16}$$

## 2.2 Vanilla PINN Training Challenges

The loss function of vanilla PINN consists of two terms: the boundary loss and the PDE loss. The boundary loss measures how well the model matches the ground truth values of the boundary condition, while the PDE loss measures how well the model satisfies the derivative values that form the PDE equation. During training, the neural network's weights are updated iteratively to minimize the loss function. This iterative update allows for the calculation of the desired outputs within the given domain. However, converging to a global minimum can be challenging because these two terms in the loss function have different scales and magnitudes. This disparity makes the loss function imbalanced and difficult to optimize.

To investigate factors that lead to an imbalance in the vanilla PINN loss function, the first benchmark is utilized. Note that vanilla PINN in this case takes x and y as inputs and outputs $u$, $v$, $p$, $\sigma_{xx}$, $\sigma_{xy}$ and $\sigma_{yy}$. Latin hypercube sampling (LHS) and random sampling methods are used to select domain points and boundary points, respectively. Figure 1 shows the distribution of the selected points. The density of selected points around the cylinder is higher than other parts of the domain to capture the underlying physics more accurately.

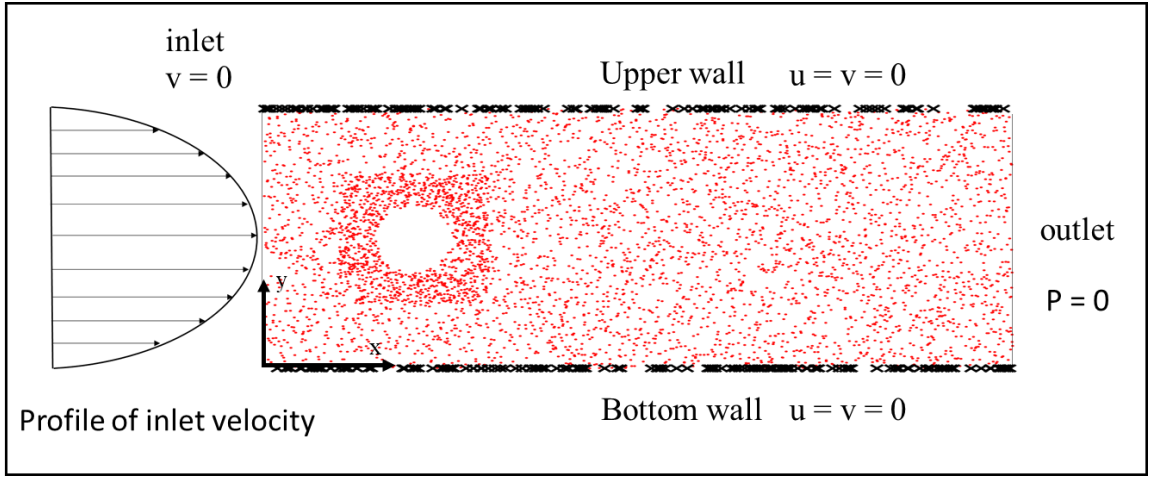

Figure 1: Selected points using LHS method and random sampling method

In this case, a slight change in the ratio of the number of domain points to boundary points from 0.035 in Case 1 to 0.033 in Case 2 leads to a change in the optimal value of $\lambda$. To quantify this impact, the vanilla PINN is trained for different values of $\lambda$ for each case, and the average training time is reported in Table 1. This table illustrates how different ratios can influence the optimal value of $\lambda$. As shown in Table 1, a minor adjustment in the ratio of the number of domain points to boundary points can result in a significant change in training time or even cause divergence. For instance, when $\lambda$ is 0.5, the average training time is considerably faster in Case 2, whereas the training time for $\lambda = 1$ is quicker in Case 1. Note that the training process is stopped when the loss value reaches a threshold of $10^{-3}$.

Table 1: Training time, in minutes, for different $\lambda$ to converge to a loss value of $10^{-3}$ in case 1 and case 2

|        | $\lambda = 0.1$ | $\lambda = 0.5$ | $\lambda = 1$ | $\lambda = 1.1$ | $\lambda = 1.5$ | $\lambda = 5$ |
|--------|---|---|---|---|---|---|
| Case 1 | Not Converged | 40.883 | 17.900 | 15.283 | 25.050 | 14.266 |
| Case 2 | Not Converged | 13.450 | 27.550 | 25.583 | 15.933 | 28.616 |

From Table 1, three factors can be identified that potentially cause the loss function to be imbalanced.

1. The first cause is related to the initial weights state of the neural network. This is a random factor and cannot be controlled or predicted, which is why the average training time is reported in Table 1. The randomness of the initial weights can lead to different learning paths during the training process, and consequently, to different local minima in the loss function.

2. As indicated in Table 1, the ratio of domain points to boundary points is another factor that can cause an imbalance in the loss function. This can be interpreted from the loss function formula itself. If more domain points are used in the loss function, the PDE loss can become dominant compared to the boundary condition.

3. The final factor, as outlined in Table 1, is the $\lambda$ value. This factor is particularly beneficial for researchers as it can be directly set in the loss function. The $\lambda$ value directly controls the value of the boundary condition loss term in the loss function. This direct control provides researchers with a powerful tool for balancing the loss function. However, it is important to note that the incorrect selection of this hyperparameter can exacerbate the imbalance in the loss function. For instance, as shown in Table 1, using a $\lambda$ value of 0.1 failed to converge to the threshold. Therefore, careful selection and tuning of the $\lambda$ value is also a time-consuming process.

While the tuning of the $\lambda$ value can be beneficial in optimizing the loss function, but it can become more challenging when a lower threshold, such as $10^{-4}$, is set to stop the training. As a result, the process becomes even more challenging and time-consuming, with an increased risk that the loss function may not converge to the desired threshold. Figure 2 illustrates how the different $\lambda$ values can affect the time of convergence and whether convergence is possible for the set threshold. As shown in Figure 2, using a value of $\lambda = 1$ makes it impossible to converge to a threshold of $10^{-4}$.

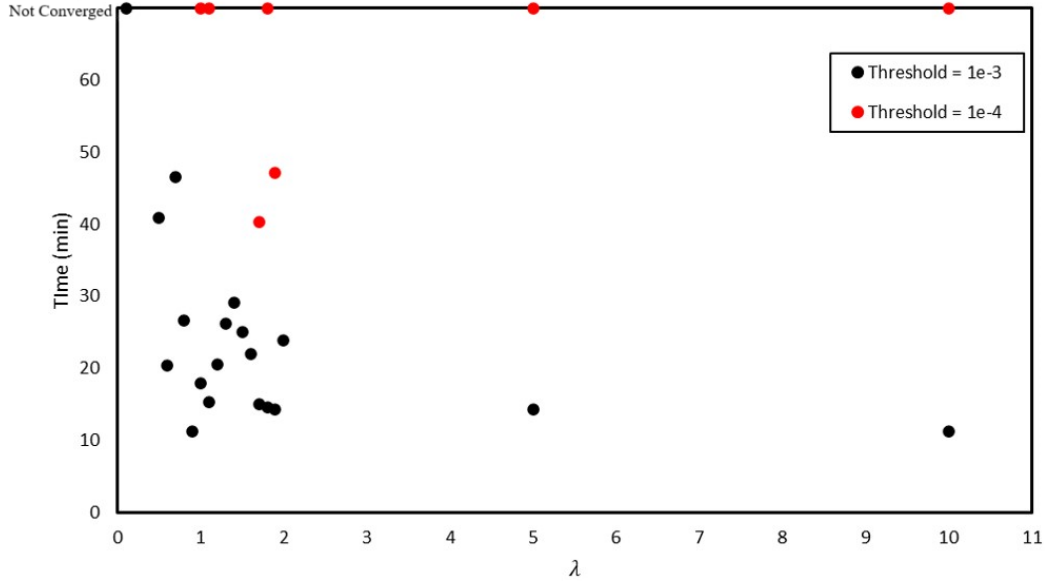

Figure 2: Illustration of how tuning $\lambda$ can impact the ability to reach different thresholds of $10^{-3}$ and $10^{-4}$ for a constant ratio of domain point to boundary point.

## 3. Methodology

As outlined in [Section 2.2](), imbalance loss function can complicate the training process. We propose that incorporating weights and biases, which are trained to learn boundary conditions, into the neural network can balance the loss function during training. This proposition is based on the premise that such initialization could steer the network toward solutions that align with the subset of boundary conditions. The logic behind this is that if a proportion of weights and biases of a neural network is capable of predicting the boundary conditions, then this proportion will influence the network's predictions. This suggests that the structure of the neural network is designed to favor the boundary conditions, which in turn influences the overall convergence of the loss function. As a result, the loss function becomes more balanced, as the network is inherently inclined to align with the boundary conditions.

In this study, our approach is named 'Feature-Enforcing-PINN' for brevity and clarity, and the term 'smart weights' is used for weights and biases informed by boundary conditions. The process of incorporating smart weights into the neural network is termed 'smart initialization'. Our research methodology's contributions are organized as follows: In [Section 3.1](), we introduce the Primary Loss Function. This function is specifically designed to guide the neural network to learn only a subset of the boundary conditions, rather than the exact solution of the PDE. Following this, in Section 3.2, we discuss the crucial role of variance reduction prior to training the neural network on the Primary Loss Function. This step is essential to ensure the stability and efficiency of the learning process. In Section 3.3, we present our new strategy for selecting domain points. Training the neural network with the Primary Loss Function and the aforementioned preprocessing steps results in what we refer to as 'smart weights'. Subsequently, we augment the network structure by adding extra layers with random weights. Finally, the enhanced neural network is trained on the complete loss function. This comprehensive approach ensures that our neural network's loss function is balanced and remains unaffected by the three primary causes discussed in Section 2.2, which typically render the vanilla PINN loss function imbalanced. Finally, in Section 4, we conduct a comparative analysis between our proposed methodology, which we refer to as FE-PINN, and Vanilla PINN.

### 3.1 Loss Function for the Smart Weights

In the process of solving PDEs, the absence of the necessary boundary conditions often leads to non-unique solutions, thereby opening up a spectrum of potential answers. Drawing from this analogy, we have devised a loss function, referred to as the Primary Loss Function, with the goal of finding one of the non-unique solutions. Training a neural network with this loss function results in a set of parameters, which we termed smart weights in Section 3. Note that smart weights are used to replace a proportion of random weights in the structure of a neural network in the training process of Vanilla PINN. The Primary Loss Function is defined as follow for each benchmark:

First Benchmark = no-slip boundary condition on side walls + inlet velocity boundary condition + $L_{pde}$ (17)

Second Benchmark = no-slip boundary condition on side walls + $L_{Inverse\ problem} + L_{pde}$ (18)

Third Benchmark = $u(x = 0, t) + L_{pde}$ (19)

Please note that the loss function defined in this section is solely for the creation of smart weights. This is distinct from the vanilla PINN loss function outlined in Section 2.1

### 3.2 Reducing Variance

When the neural network undergoes training with both smart and random weights, the smart weights begin to lose their information about the boundary conditions after a few iterations if they are initialized similarly to random weights. This is attributed to the similar update rates of both smart and random weights, which result from their comparable gradients and identical learning rates. Consequently, the training process becomes imbalanced, resembling the behavior of a vanilla PINN, as discussed in [Section 2.2](#).

To prevent smart weights from losing their information, a novel initialization approach is used for the process of creating smart weights before the first phase of training. In this approach, random weights are initialized with a lower variance, which leads to smaller initial gradients. After reducing variance and training the neural network with the Primary Loss Function, the newly created smart weights are utilized to replace a portion of the random weights within the structure of the neural network. These random weights, which are initialized using the Xavier initialization method, exhibit a higher variance than smart weights and consequently, larger gradients. As a result, they tend to update more rapidly than the smart weights during the training process on the complete loss function from [Section 2.1](#) for each benchmark. Given that the smart weights are predominantly identical to each other and possess smaller gradients, they undergo minimal changes during the training process on the loss function of a vanilla PINN. This novel strategy ensures that the smart weights maintain their knowledge about the boundary conditions for a greater number of iterations due to their lower variance, smaller gradients, and identical response to input. To elucidate the influence of variance reduction on the gradient and $L_{pde}$ before the training process, we present [Table 2](#) and [Figure 3](#). [Figure 3,](#) in particular, compares the outputs and their derivatives with respect to inputs after reducing the initialization variance and a Xavier-initialized network. For instance, in the first benchmark, output values (u, v, p) after reducing the variance are near zero and exhibit minimal changes, while those for the Xavier-initialized network vary more significantly as illustrated in [Figure 3(a & b)](#). Also note that the derivative values after reducing variance are much smaller than those of the Xavier-initialized network, indicating smoother variation across the domain as shown in [Figures 3(c & d).](#) This observation is confirmed by repeating the process with different random seeds. [Table 2](#) presents the $L_{pde}$ for both networks. The network with lower variance assigns a lower PDE loss value due to its smaller derivative values as reported in [Table 2](#). This pattern holds across different random seeds and domain points, suggesting that the procedure is not based on randomness but is a systematic effect of variance reduction. As the final point, the impact of variance reduction on smart weights during the second phase of training is investigated. Empirical results indicate a trade-off between not learning the Primary Loss Function and information loss when variance reduction is implemented as a preprocessing step before the initial training phase. This suggests that if the variance is reduced by a small factor, there is no significant impact on the convergence process. However, if the reduction factor is large, convergence does not occur. Empirical evidence recommends setting the variance reduction factor within the range of $\sqrt{5}$ to $\sqrt{10}$. It's noteworthy that the first two benchmarks utilized $\sqrt{10}$, while the final benchmark employed $\sqrt{5}$. Note that, the variance reduction factors differences between the benchmarks are attributed to the complexity of the neural networks used in each benchmark.

Table 2: Calculated PDE loss for the untrained averaged-out neural network and a neural network initialized with Xavier scheme

|  | Total domain points | Xavier-initialized network ($L_{pde}$) | Network with reduced variance ($L_{pde}$) |
| --- | --- | --- | --- |
| random seed #1 | 371760 | 1.2691 | 0.0034 |
| random seed #2 | 371760 | 1.4661 | 0.0049 |
| random seed #3 | 371760 | 0.9479 | 0.0007 |
| random seed #4 | 371760 | 0.5480 | 0.0043 |

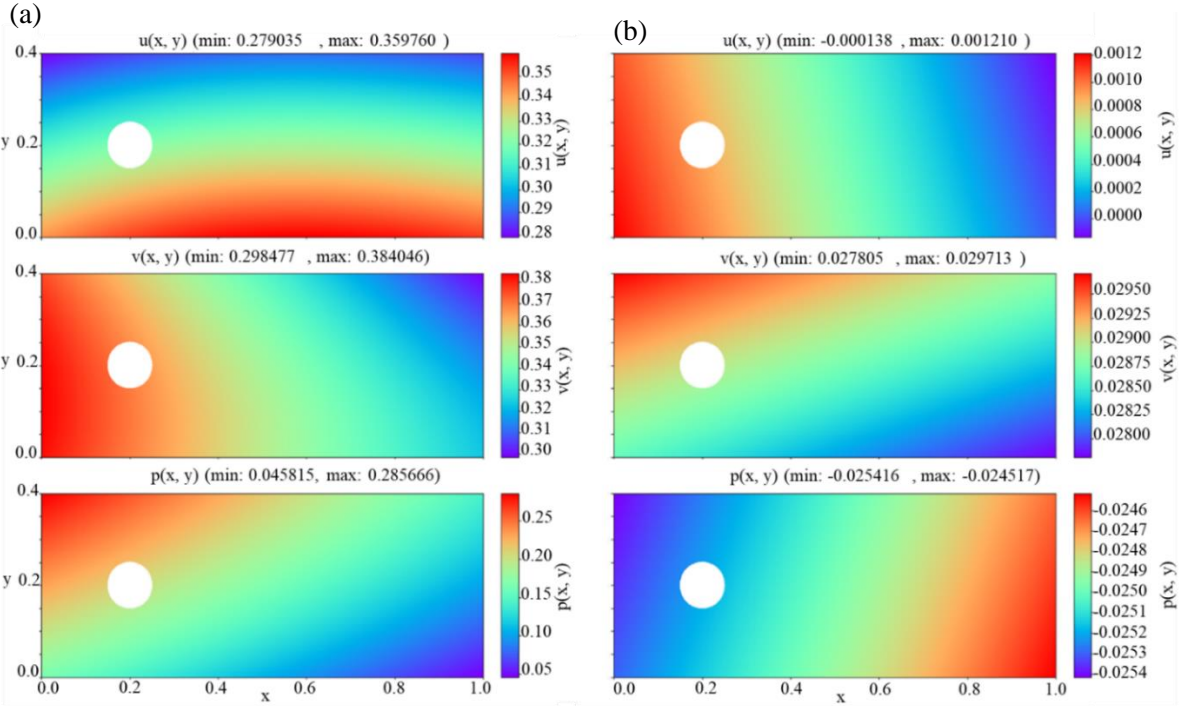

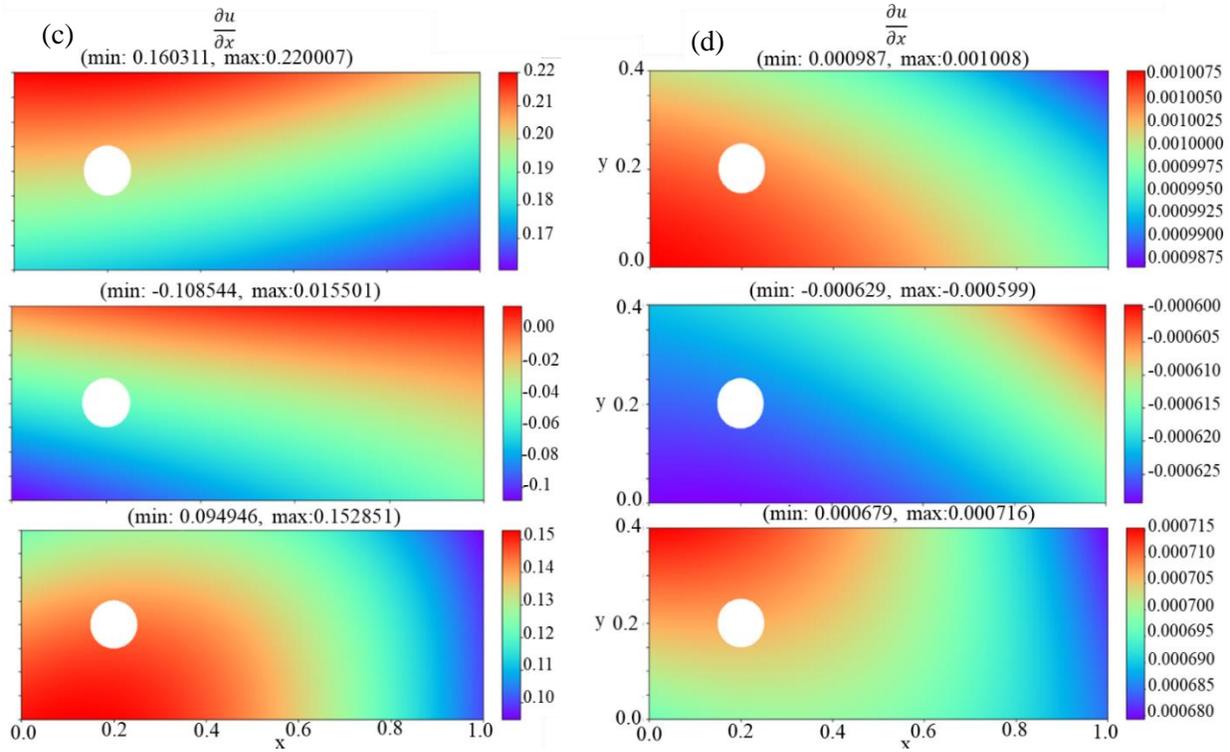

Figure 3: The first set of three figures, labeled as (b), illustrate the variables (u, v, p) for the neural network post variance reduction. In contrast, figures labeled as (a) represent the neural network initialized using the Xavier scheme. The final set of three figures exhibit the value of du/dx for two scenarios: one for the neural network after variance reduction (d), and the other for a neural network initialized using the Xavier scheme(c).

### 3.3 Two-Phase Training

An alternative strategy for LHS involves the computation of PDE residuals at each iteration of training the neural network. If these residuals exceed a predetermined threshold for domain points, they are chosen as input points for that iteration. This iterative method, introduced by Arka Daw et al., is feasible because PDE residuals are a function of a neural network's weight state [43]. The Xavier initialization method, when applied with different random seeds, results in different weight configurations for a neural network. As the loss function is dependent on the weight state, this leads to diverse outcomes when calculating the PDE residual for domain points. Utilizing different random seeds and selecting domain points that surpass a specific PDE residual threshold result in a unique set of domain points for each initial weight state. Consequently, with each change in the random seed, the focus shifts to a different part of the domain, from which the domain points are then selected. This results in a higher density of points in that area compared to the rest of the domain. This uneven distribution leads to the exclusion of certain parts of the domain from the selection of domain points. This exclusion allows the neural network to concentrate on its interpolation capabilities within the excluded areas. To clarify, the process we utilize for the selection of domain points is designed with an emphasis on using interpolation across the excluded parts of the domain. In other words, rather than learning the exact solution, the network is trained to make predictions within certain boundaries. This is due to the presence of only a subset of boundary conditions in the Primary Loss Function. This procedure, which employs N different random seeds to select domain points, is executed before the training process begins. This strategy for domain point selection is a modification of the method introduced by Arka Daw. Note that N is a hyperparameter that determines the number of different random seeds used for initializing the neural network's weights. The choice of N can influence the diversity of the domain points selected during training. However, for a wide range of values, the outcome remains consistent, indicating that the exact value of N is not critical to the strategy's success. Empirical observations suggest that using between 5 to 10 random seeds yields similar results.

This novel approach is used to select domain points for all the benchmarks in the first training phase. We train a neural network on the Primary Loss Function for each benchmark using the selected domain points. This training occurs after the preprocessing step outlined in [Section 3.2](). All benchmarks are trained exclusively using a GPU RTX 4080. Following the initial training phase, smart weights are produced. Upon the generation of smart weights for each benchmark, these weights are subsequently utilized to substitute a proportion of the weights within a neural network, which was initialized via the Xavier initialization method. In the training process of a neural network, we strategically replace the weights in both the output layer and the initial few layers with smart weights. This replacement ensures that these critical layers have a more pronounced impact on the learning process and the overall performance of the network. It's important to note in a neural network these layers have distinct roles. The output layer, for instance, is crucial as it is directly responsible for the final prediction made by the neural network. Conversely, the initial layers are typically tasked with learning the basic features of the input data, serving as the foundation for the subsequent layers. This strategic weight replacement enhances convergence and balances the training for the second phase. After the addition of smart weights, the second training process is carried out on the loss function, defined in [Section 2.1](). Unlike the first training phase, the LHS method is used to select domain points for this phase. Figure 4 depicts the structure of the neural network for the first two benchmarks. In this figure, the smart weights, which are obtained after training on the Primary Loss Function, are represented by blue lines. Conversely, the red lines illustrate the random weights that were added using Xavier initialization. It is imperative to acknowledge that the initial two benchmarks utilize 'x' and 'y' as inputs, whereas the final benchmark employs 'x' and 'time'. The neural network for the first two benchmarks comprises eight hidden layers, each encompassing 40 neurons. The weights of half of these hidden layers, precisely four, are substituted with smart weights. Conversely, the neural network for the final benchmark consists of four hidden layers, with the weights of half of these layers, precisely two, being replaced with smart weights. It is noteworthy that the increased complexity of the first two benchmarks, in comparison to the last one, necessitates a greater number of hidden layers in these two benchmarks.

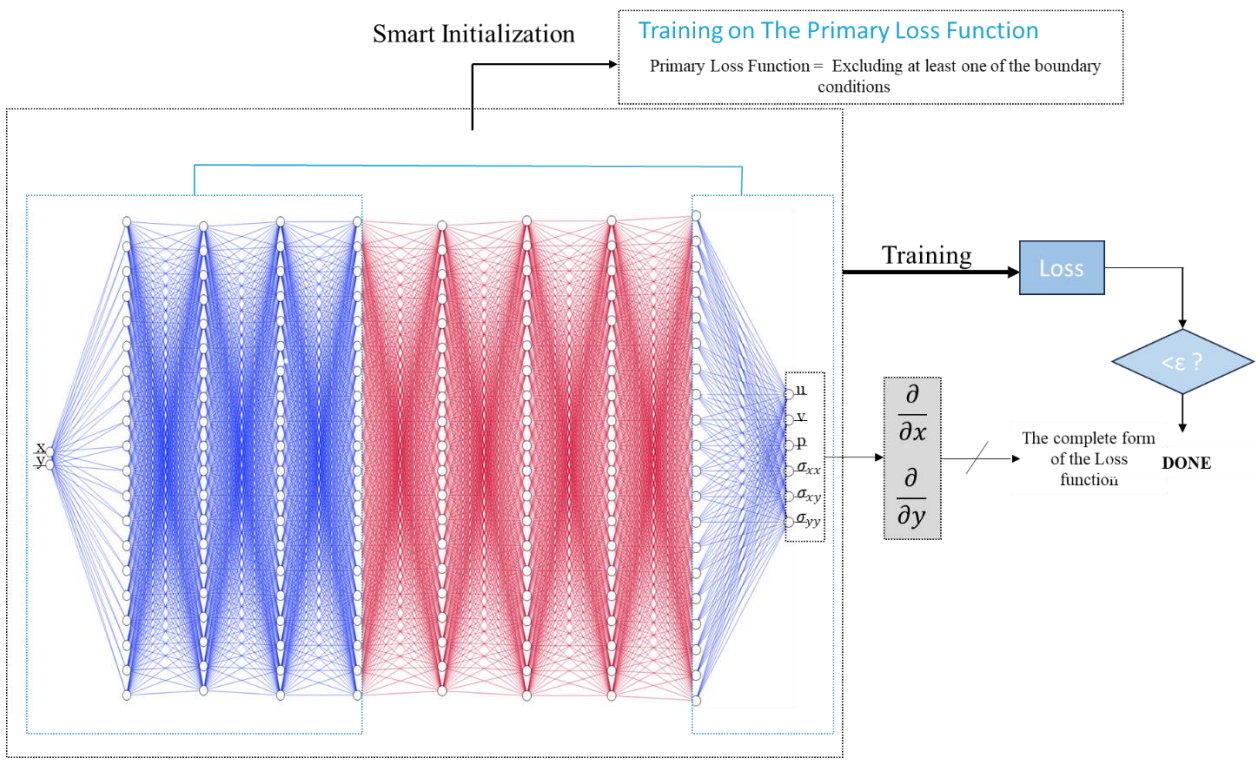

Figure 4: A schematic of the FE-PINN structure

## 4 Results and Discussion

### 4.1 First Benchmark

In the training process, the ADAM optimizer is used with a learning rate of $3\times10^{-4}$, initially focusing on the Primary Loss Function. This training enables the network, which has been trained on the Primary Loss Function (Equation (17)), to predict the values of u, v, and p, as shown in Figure 5. This figure illustrates how smart weights predict outputs for the first benchmark. It's worth noting that the average training time for the Primary Loss Function in this benchmark is 1.2 minutes. For enhanced visualization in Figure 5, the trained networks are provided with a comprehensive set of domain points, especially the points surrounding the cylinder. Upon completion of the training, smart weights are generated. It is noteworthy that the training process is cost-effective and fast.

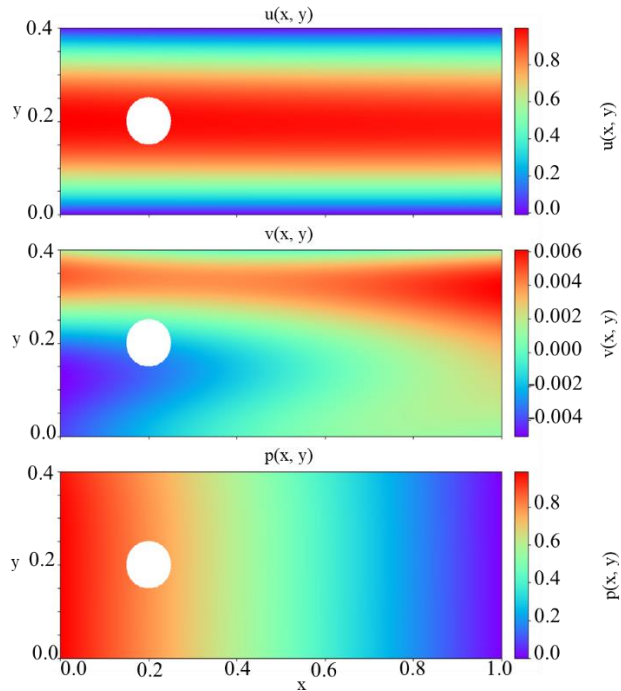

Figure 5: how smart weights predict *u, v, p* for the first benchmark

Upon the creation of smart weights and the enhancement of the network's complexity by appending additional layers atop them before the output layer, the subsequent step entails training the newly established model. The aim is to augment the accuracy of its predictions on the complete loss function (Equation (14)) using random weights. In the second training process, the parameter $\lambda$ is being kept at 1 for the entire training process for our approach while tuned for vanilla PINN. Domain points are selected utilizing the LHS method, as depicted in Figure 1. The optimization process employs the LBFGS method from the Torch library. The smart initialization introduced in this study can supplant the time-consuming process of $\lambda$ tuning. As shown in Table 3, on average, the smart initialization and training times are faster compared to the hyperparameter tuning and training times of the vanilla PINN, respectively. We employ a random search to quantify the average time required to find the optimal $\lambda$ value for the vanilla PINN. This process is repeated five times to ensure the total time is independent of randomness. For each iteration, the tuning time and the convergence time of the best $\lambda$ value are measured. The average times are then calculated and reported in Table 3. On average, the smart initialization process in FE-PINN is 144 times faster when using the Primary Loss Function (Equation (17)) instead of tuning $\lambda$ directly. FE-PINN outperforms the vanilla PINN by accelerating the average training time by 2.2 times. This demonstrates that our approach not only eliminates the need for hyperparameter tuning but also trains faster on the target task.

Table 3: Time on average, in minutes, required for tuning time, smart initialization and training time for vanilla PINN and FE-PINN

|  | Average Tuning time | Smart Initialization | Average training time | Total time for reaching $10^{-4}$ |
| --- | --- | --- | --- | --- |
| Vanilla PINN | 172.3 | ----- | 47.4 | 219.6 |
| FE-PINN | ----- | 1.2 | 21.4 | 22.6 |

The Big O notation is utilized to characterize the growth rate of the number of operations, disregarding constant factors and lower-order terms. The Big O (time complexity) of training a vanilla PINN is contingent on various factors such as the total number of layers, the total number of neurons per layer, the number of input and output features, the number of domain points, the number of epochs, calculated derivatives, the optimization algorithm, and the tuning of the $\lambda$ value. In Table 3, both FE-PINN and the vanilla PINN share identical characteristics, with one distinction: the $\lambda$ value in the vanilla PINN necessitates tuning, but our approach substitutes this process with a low-cost, smart initialization process. The difference between the Big O notation for FE-PINN and the vanilla PINN lies in the necessity to explore the optimal value for $\lambda$ in the loss function of the vanilla PINN. For the vanilla PINN, in the best-case scenario, the first selected $\lambda$ value converges to the desired threshold (O(1)), while in the worst-case scenario, all possible values must be explored (O(n)), where n is the number of iterations performed to find the best value of $\lambda$. However, in the case of our approach, there is no need for such a process. Instead, a low-cost smart initialization can replace the exploration of different $\lambda$ values, with an average training time of just one minute. It is important to note that for the FE-PINN, the time complexity is O(1), which means it is constant and does not depend on finding the optimum $\lambda$ value.

In our investigation of the other two factors outlined in Section 2.2, we refer to Table 4. It's crucial to note that the initial weight state of both the vanilla PINN and FE-PINN share the same weights in the red layers of Figure 4 for each row of Table 4. The key difference lies in the blue layers, which are created using a smart initialization process for FE-PINN, while Xavier initialization is used for the vanilla PINN. Smart weights as blue layers for our model make the loss function balanced while random weights as blue layers make the vanilla PINN loss function imbalanced. The time required for smart initialization is detailed in the second column of Table 4, represented by the first number before the addition sign, and the second number is the training time of FE-PINN. In order to address the second cause mentioned in Section 2.2., we report the time required for convergence, considering various ratios of domain points to boundary points. Each ratio is reported three times to account for the potential influence of the initial weight state on the convergence process which is the first cause. For instance, at a ratio of 41.7, FE-PINN converges to the desired threshold, which is $10^{-4}$, but the vanilla PINN cannot converge in two of the three cases. The reasons behind the inability of the vanilla PINN to converge are the first two causes mentioned earlier. However, in certain cases, such as ratios of 43.81 and 20.72, the vanilla PINN fails to converge to the desired threshold with three different initial weight states. This suggests that the imbalance is caused by the ratio of domain points to boundary points in these instances. In contrast, our approach ensures that the loss function remains balanced across different ratios and converges in all cases of Table 4. Also, in some ratios where the initial weight state causes the vanilla PINN to fail to converge, FE-PINN still converges. For instance, in the first case of ratio 34.76 with the same initial weight state in the red layers (Figure 4) for both models, FE-PINN converges in 33.4 minutes while the vanilla PINN fails to converge. In the second case, when the vanilla PINN manages to converge, our model still converges 1.5 times faster. It's important to note that in cases where vanilla PINN converges, not only does our model converge, but it also converges faster than the vanilla PINN. Tables 3 and 4 demonstrate the fact that the loss function is balanced for all three causes mentioned in Section 2.2, highlighting a key feature of FE-PINN and its robust performance across various scenarios.

Table 4: The training time, measured in minutes, is given for both vanilla PINN and FE-PINN under different ratios and initial weight states.

|        | ratio  | FE-PINN   | Vanilla PINN  |
|--------|--------|-----------|---------------|
| Case 1 | 41.7   | 1.2+20.4  | Not Converged |
| Case 2 | 41.7   | 1.2 +21.1 | Not Converged |
| Case 3 | 41.7   | 1.0+23.3  | 31.1          |
| Case 1 | 38.28  | 1.1+24.0  | 29.1          |
| Case 2 | 38.28  | 1.1+28.0  | 32.7          |
| Case 3 | 38.28  | 1.1+17.3  | Not Converged |
| Case 1 | 45.23  | 1.1+24.0  | Not Converged |
| Case 2 | 45.23  | 1.3+18.0  | 26.3          |
| Case 3 | 45.23  | 1.0+22.4  | 32.5          |
| Case 1 | 43.81  | 1.3+26.5  | Not Converged |
| Case 2 | 43.81  | 1.1+17.9  | Not Converged |
| Case 3 | 43.81  | 1.2+22.0  | Not Converged |
| Case 1 | 34.76  | 1.0+32.4  | Not Converged |
| Case 2 | 34.76  | 1.3+18.2  | 30.0          |
| Case 3 | 34.76  | 1.3+28.1  | Not Converged |
| Case 1 | 20.72  | 1.2+21.5  | Not Converged |
| Case 2 | 20.72  | 1.2+26.0  | Not Converged |
| Case 3 | 20.72  | 1.3+24.1  | Not Converged |

Figure 6 offers a straightforward depiction of how the loss function of our model reaches its global minimum after the second phase of training smoothly. This highlights the success of the training process in optimizing the loss function and achieving the desired outcome. As shown in this figure, the loss function decreases drastically in the initial iterations. Also, note that the magnified figure shows that the loss value converges to an order of magnitude of $10^{-5}$. In Figure 7, the outputs of our methodology and the vanilla PINN are compared visually. Note the difference between Figure 7 and Figure 5. Figure 5 is the prediction of our model after the first training phase while Figure 7 is the result the second training phase. Finally, the FE-PINN predictions are validated using simulation results, keeping the *'y'* constant for velocity magnitude in Figure 8.

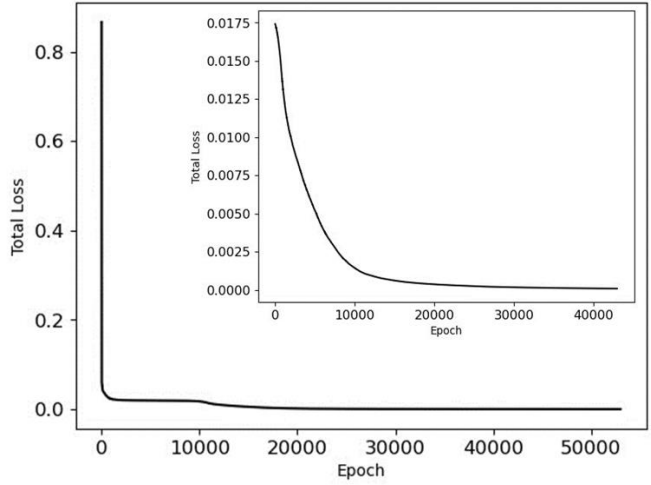

Figure 6: Illustration of how the neural network total loss value changes versus Epoch.

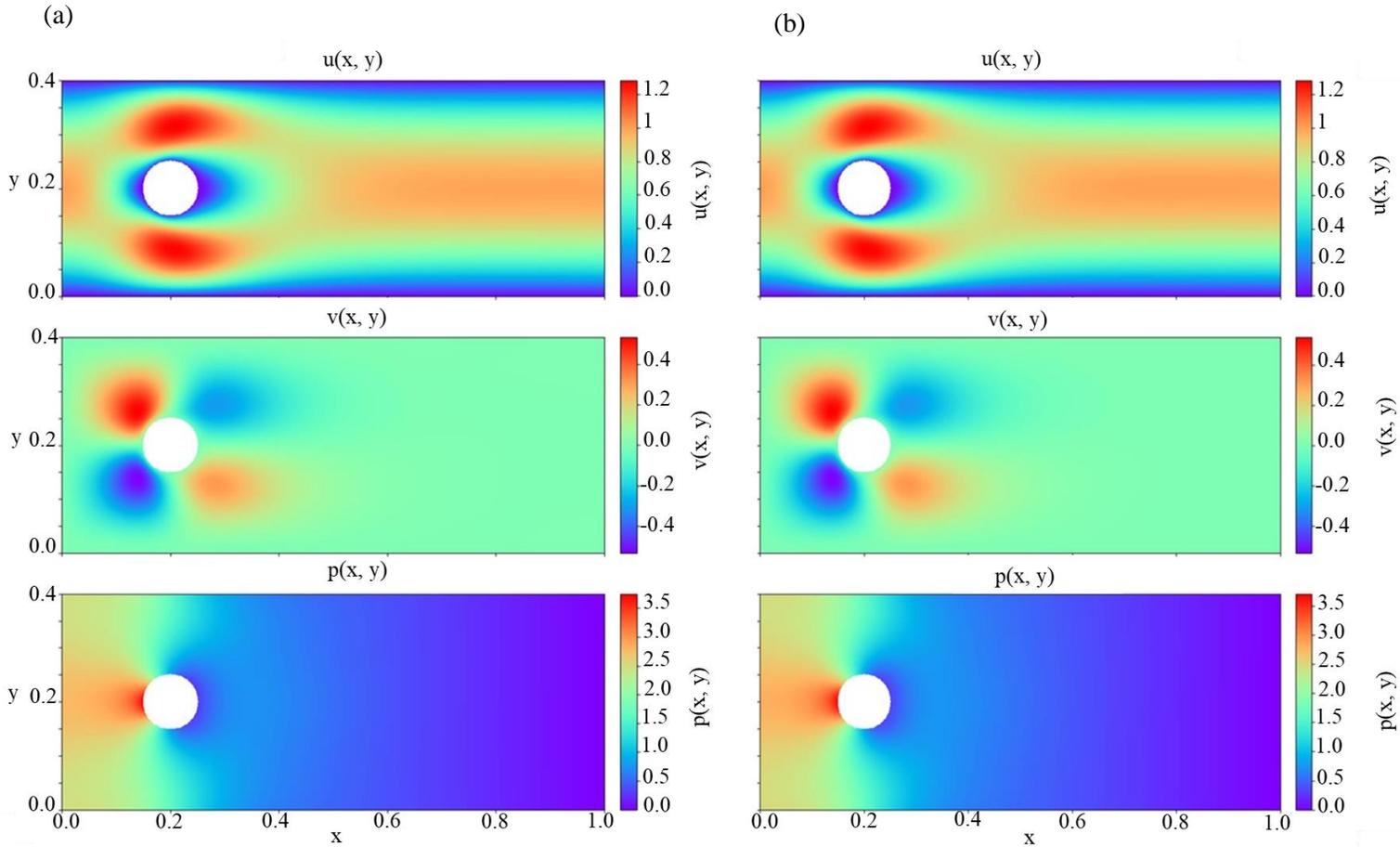

Figure 7: A comparison is made between the (u,v,p) predicted by (a) FE-PINN and (b) vanilla PINN for the flow over a 2D cylinder.

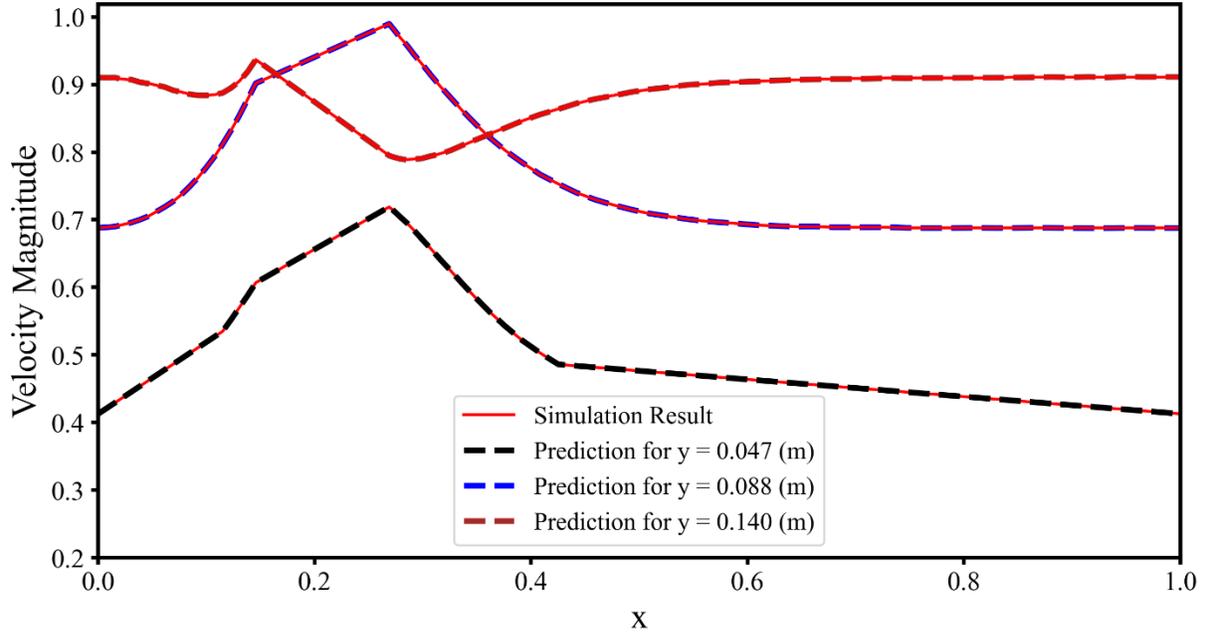

Figure 8: Validation results for FE-PINN

## 4.2 Second Benchmark

Vanilla PINN has shown promise in addressing inverse problems, which are inherently ill-posed due to the presence of unknown boundary conditions. In this section, we present a comparative analysis of the performance of FE-PINN and the vanilla PINN in resolving an inverse problem characterized in Section 2.1. Benchmark 2 is depicted in Figure 9. Furthermore, we have 60 domain points with known u and v values, as denoted by the blue marks in Figure 9. As previously described, the neural network is initially trained on the Primary Loss Function (Equation (18)). Following the smart initialization process introduced in this study, the model is then trained on Equation (15) to solve the problem.

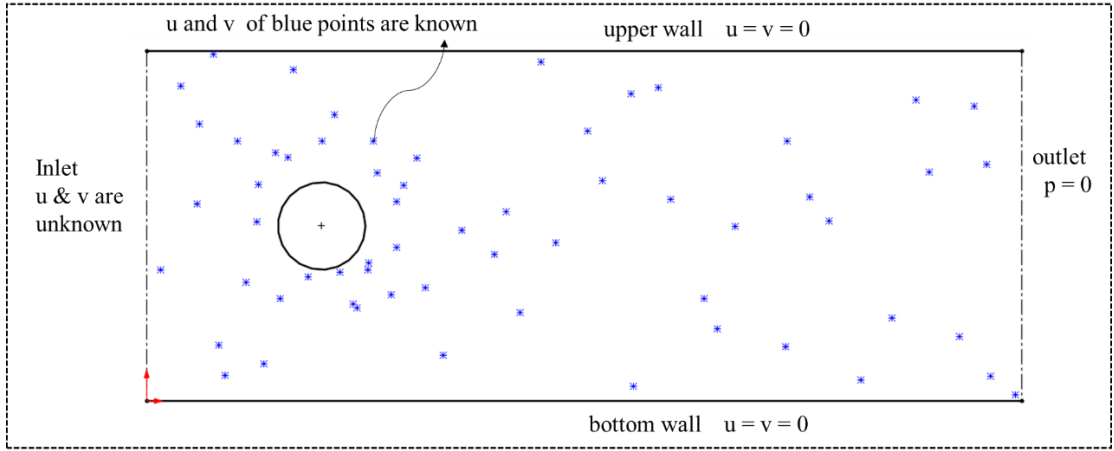

Figure 9: Inverse Problem of finding inlet velocity.

In order to ensure a fair comparison, both FE-PINN and the vanilla PINN are trained under comparable conditions. This includes employing the same learning rate, the same optimization algorithm (LBFGS), an identical number of domain and boundary points, and a similar structure. It is important to note that the structure of both the vanilla PINN and our approach remains consistent with the previous section, encompassing the same number of layers, inputs, and neurons. The training phase for both methods is terminated when the loss value reaches a threshold of $10^{-4}$. Table 5 provides a comparison of the average training time of FE-PINN and the vanilla PINN. To ensure that the total time is independent of the first cause of imbalance mentioned in Section 2.2, FE-PINN is trained three different times. The other columns of Table 5 display the average training time for the vanilla PINN with different $\lambda$ values. The final row of Table 5 employs the R-squared metric to compare the efficacy of these models in predicting the inlet velocity. This metric provides a measure of how well the predicted values fit the actual values, thereby offering a quantitative assessment of the model's performance.

Table 5: Displaying the time, in minutes, required to converge to a total loss value of $10^{-4}$.

|  | $\lambda = 1$ | $\lambda = 1.3$ | $\lambda = 1.5$ | FE-PINN #1 | FE-PINN #2 | FE-PINN #3 |
|---|---|---|---|---|---|---|
| Smart Initialization | -------- | -------- | -------- | 0.99 | 1.2 | 1.3 |
| Training Time | 24.65 | 27.86 | 35.40 | 12.99 | 13.11 | 13.61 |
| Accuracy | 0.9980 | 0.9975 | 0.9960 | 0.9983 | 0.9983 | 0.9980 |

As demonstrated in Table 5, the method proposed in this study is approximately twice as fast as the vanilla PINN in this benchmark, eliminating the need for tuning the $\lambda$ value. While tuning the $\lambda$ value can expedite the convergence for vanilla PINN, it is noteworthy that in the case of the inverse problem examined in this study, most of the domain to boundary points ratios for vanilla PINN can reach the threshold of $10^{-4}$, with a few exceptions.

**4.3 Third Benchmark**

The final benchmark under consideration necessitates fewer derivatives to be computed and has fewer boundary conditions compared to the first two benchmarks. The decrease in terms within the loss function results in a more balanced loss function relative to the first benchmark. Consequently, the loss function is less influenced by the three primary sources of imbalance, as outlined in Section 2.2. To examine the influence of the initialization with random weights, the ratio of the domain to boundary points, and $\lambda$ value on the balance of the loss function for this benchmark, we refer to Table 6. In the given table, each row, such as the one represented by Table 4, maintains the same random weights across all models in each column. However, there is a distinct difference when it comes to the first hidden layer and the output layer for the 'FE-PINN' column. Random weights in these layers are replaced by smart weights. Note that for every ratio, each of the three cases uses a unique random seed. This accounts for the influence of the initial weight state on the results. In the third column of the table, the first number signifies the outcome of the smart initialization process. This process involves training the neural network on the Primary Loss Function specific to this benchmark (Equation(19)). The second number, which follows the addition sign, represents the training time. All times for this benchmark are reported in seconds. The training process is designed to stop once the total loss value reaches a threshold of $1\times10^{-6}$. When this threshold is reached, the neural network is able to accurately predict the solution, as illustrated in Figure 10. Upon closer examination of Table 6, it becomes apparent that our proposed method exhibits a faster convergence rate compared to the vanilla PINN across all considered ratios in each case. For the vanilla PINN to achieve a training time comparable to our method, the three main factors that influence the loss function must be in their optimal states. For instance, in Case 3, where the ratio is 11.25, all three factors are near their optimal states for the vanilla PINN with a $\lambda$ value of 1. Consequently, the training time of the vanilla PINN is comparable to the sum of our approach's training time and its smart initialization process in this case. In contrast, in case 1 from the previous ratio, the only altered factor is a new random weight state. Note that in this case, the training time is different from case 3 as expected since the randomness of the initial weights can lead to different learning paths during the training process. However, in this case, FE-PINN converges 1.5 times faster than the vanilla PINN. This suggests that, despite the change in random weights, the loss function remains balanced due to the presence of smart weights. However, in the case of vanilla PINN, this change in random weights results in a more imbalanced loss function, which is the reason for its longer training time compared to FE-PINN. It is important to note that finding an initial state that results in a balanced loss function is nearly impossible in the vanilla PINN as it is entirely dependent on randomness. In our approach, the smart weights neutralize the effect of the initial state on the loss function. This demonstrates the robustness of our approach in maintaining balance and achieving faster convergence, regardless of the randomness introduced by different seeds. Furthermore, FE-PINN consistently outperforms the vanilla PINN across all ratios, indicating that its loss function is highly balanced and less affected by different causes. Notably, this balance is maintained across an extensive range of ratios, demonstrating the method's adaptability and robustness. This characteristic is particularly significant as it negates the necessity to identify an optimum ratio for convergence. Lastly, while the $\lambda$ factor can be tuned for the vanilla PINN to match the convergence speed of our approach, this process is time-consuming compared to the process of smart initialization, which in this case took only about 5 seconds. This is significantly faster when compared to the time-consuming task of tuning the $\lambda$ factor for this benchmark. Therefore, our approach not only addresses all causes that make loss function imbalance but also reduces the training time. It's important to highlight that the last benchmark is less complex than the initial one. Consequently, it has a higher probability of converging to the desired threshold, primarily due to a more balanced loss function. Our methodology demonstrates efficacy even in the context of simpler benchmarks. However, the effectiveness of our approach becomes more evident when tackling more complex problems where the loss function is significantly influenced by imbalances, as observed in the first benchmark.

Table 6: Displaying the time, in seconds, required to converge to a total loss value of $10^{-6}$.

| | Ratio | FE-PINN | Vanilla PINN | | | | |
|---|---|---|---|---|---|---|---|
| | | | $\lambda = 1$ | $\lambda = 1.2$ | $\lambda = 1.4$ | $\lambda = 1.6$ | $\lambda = 1.8$ |
| Case 1 | 6.25 | 6+46 | 81 | 86 | 92 | 95 | 110 |
| Case 2 | 6.25 | 4+32 | 68 | 65 | 72 | 72 | 80 |
| Case 3 | 6.25 | 4+42 | 71 | 82 | 82 | 83 | 88 |
| Case 1 | 8.75 | 4+46 | 72 | 64 | 67 | 74 | 79 |
| Case 2 | 8.75 | 6+61 | 72 | 78 | 88 | 82 | 73 |
| Case 3 | 8.75 | 6+35 | 47 | 55 | 57 | 56 | 62 |
| Case 1 | 11.25 | 5+34 | 59 | 57 | 60 | 61 | 64 |
| Case 2 | 11.25 | 6+40 | 79 | 89 | 92 | 92 | 91 |
| Case 3 | 11.25 | 5+61 | 67 | 83 | 96 | 95 | 96 |
| Case 1 | 15 | 5+49 | 93 | 98 | 106 | 121 | 123 |
| Case 2 | 15 | 5+60 | 72 | 89 | 92 | 97 | 101 |
| Case 3 | 15 | 5+28 | 81 | 89 | 96 | 109 | 119 |
| Case 1 | 18.75 | 5+82 | 91 | 96 | 101 | 106 | 114 |
| Case 2 | 18.75 | 5+39 | 105 | 107 | 112 | 139 | 135 |
| Case 3 | 18.75 | 4+41 | 106 | 106 | 122 | 130 | 134 |

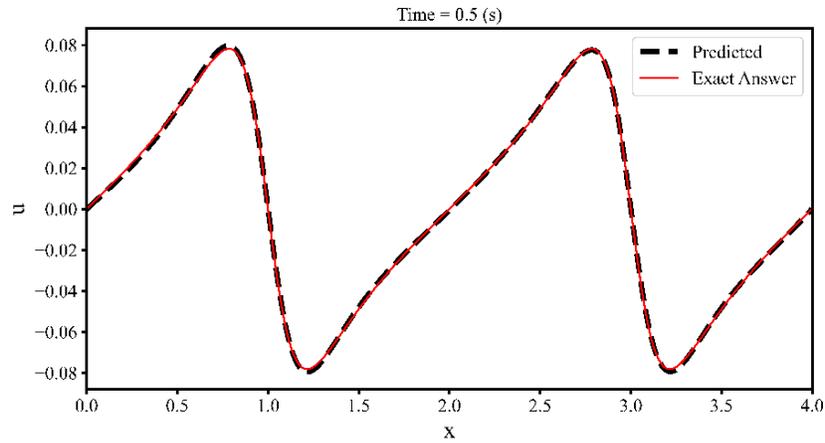

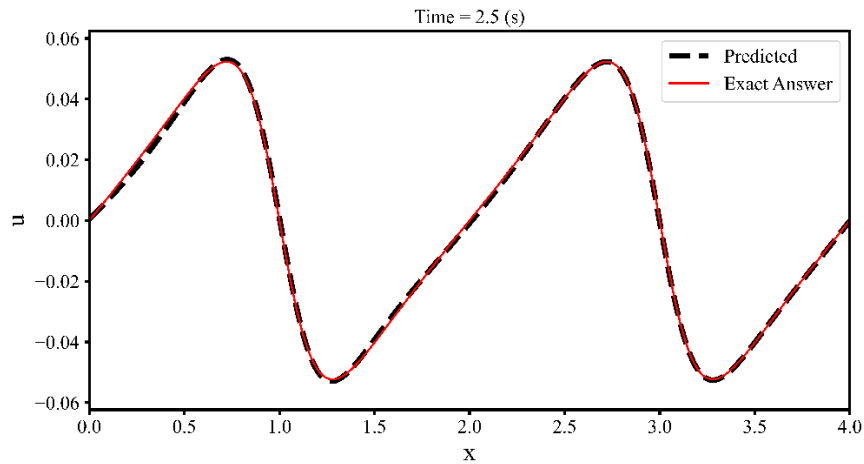

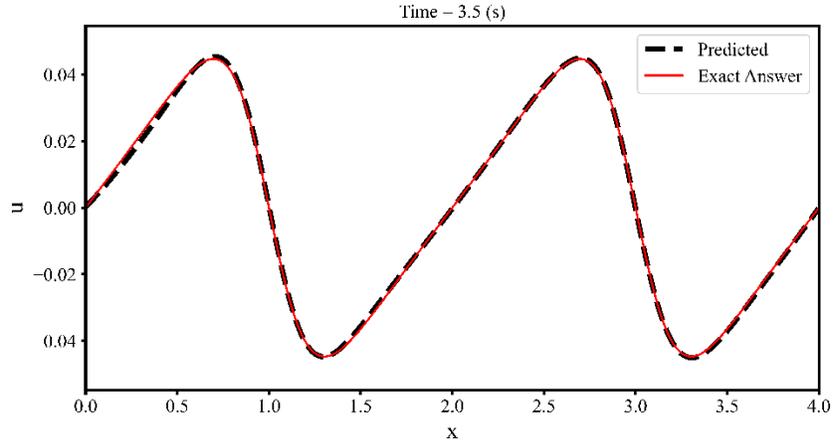

Figure 10: Validation results for FE-PINN

## 5. Conclusion

This study tackles the convergence issues of the Vanilla Physics Informed Neural Network (PINN), a method for solving partial differential equations (PDEs). The main issue with the Vanilla PINN is its struggle to converge due to an imbalanced loss function. We identified three main factors contributing to this imbalance: the initial weight state of a neural network, the ratio of the domain to boundary points, and the loss weighting factor. To tackle these challenges, we proposed a two-phase training process termed as Feature-Enforcing PINN (FE-PINN). In the first phase, we introduce a novel loss function and preprocessing steps including reducing initialization variance, and domain point selection based on different initial neural network states, producing "smart weights". The second phase is like Vanilla PINN training, but a proportion of random weights are replaced by smart weights. Three benchmarks were used to compare our approach with the Vanilla PINN. Our findings indicate that our approach converges faster than the Vanilla PINN, even when hyperparameter tuning is employed to balance the loss function. In contrast to the Vanilla PINN, the smart weights in our study neutralize the effects of the three aforementioned factors on the loss function. Our approach performs best in more complex problems, which have more boundary conditions and derivatives. In these scenarios, the Vanilla PINN fails to converge and needs to find an optimum state for the imbalance factors. However, our approach is not affected by these factors. Even in simpler problems, our approach is still faster than the Vanilla PINN. It is noteworthy that the initial phase of our training methodology exhibits superior efficiency and speed compared to the process of determining the optimal ratio and loss weighting strategies. For instance, in the first benchmark, the initial phase of training was completed in approximately one minute, which is 144 times faster than the loss weighting process in a Vanilla PINN. The approach introduced in this study effectively balances the loss function for various factors, while maintaining a faster speed. This makes it robust and suitable for a wide range of applications.

**Author Contributions:** Mahyar Jahaninasab; Methodology & writing Mohamad Ali Bijarchi; review & editing

**Funding:** This research received no external funding.

**Data Availability Statement:** Data sharing is not applicable to this article as no new data were created or analyzed in this study.

**Conflicts of Interest:** The authors declare no conflict of interest.

**Acknowledgment:** The authors would like to acknowledge Miss. Mina Rezaie & Mr. Ehsan Ghaderi for their helpful and kind support.